\documentclass[11pt]{article}

\usepackage[utf8]{inputenc}
\usepackage[T1]{fontenc}
\usepackage{lmodern}
\usepackage[margin=1in]{geometry}
\usepackage{amsmath}
\usepackage{amssymb}
\usepackage{booktabs}
\usepackage{algorithm}
\usepackage{algpseudocode}
\usepackage{hyperref}
\usepackage{enumitem}
\usepackage{tikz}
\usetikzlibrary{arrows.meta,positioning,fit,backgrounds}

\hypersetup{
    colorlinks=true,
    linkcolor=blue,
    citecolor=blue,
    urlcolor=blue
}

\title{Learning Dynamic User Personas from Implicit Interaction Streams via Iterative Refinement}
\author{Haifeng Wu \\ \texttt{haifwu@paypal.com}}
\date{}

\begin{document}

\maketitle

\begin{abstract}
Personalizing large language models (LLMs) to individual users is essential for improving long-term user experience, yet existing approaches typically rely on explicit preference supervision such as pairwise comparisons or demographic attributes, limiting their applicability in natural interaction settings. We propose IRIS, a framework that learns dynamic user personas directly from implicit interaction streams. Instead of requiring explicit feedback, IRIS leverages language models to extract behavioral signals from everyday conversations, infer latent user preferences, and continuously refine persona representations through a prediction-driven closed loop.

To evaluate dynamic personalization, we introduce a protocol based on behavior prediction, persona stability, and decision prediction. We first conduct a proof-of-concept study on a synthetic interaction stream derived from public-domain autobiographical text, demonstrating that IRIS produces stable persona representations and consistently distinguishes individual users, while also revealing the limitations of memory-only approaches on recall-oriented evaluation metrics. We then validate the framework on anonymized real-world Reddit r/AmItheAsshole (AITA) data, where personas are constructed solely from each author's historical interactions. Across 100 authors, IRIS achieves the highest decision prediction accuracy among all evaluated methods (61.0\%), outperforming static personas, memory-only retrieval, and no-personalization baselines.

Our results suggest that implicit behavioral modeling provides a scalable alternative to explicit preference learning for personalized LLMs. Beyond conversational assistants, the proposed closed-loop framework offers a practical foundation for adaptive embodied agents and long-term human--AI interaction, where continuously evolving user models are essential for anticipating preferences and supporting personalized behavior over time.
\end{abstract}

\section{Introduction}

A user who asks an AI assistant to ``clean up this email'' on Monday expects a different outcome than when the same user issues the same request on Friday before a board presentation. Tone, length, formality, and emphasis all shift --- not because the task changed, but because the \textit{person} did, at least in context. Capturing this shifting sense of self is the central challenge of LLM personalization.

Existing personalization approaches fall into two categories, and both are inadequate in practice. The first relies on \textbf{explicit preference signals}: curated pairwise comparisons, thumbs-up/thumbs-down ratings, or demographic questionnaires. These signals are expensive to collect at scale and suffer from the cold-start problem --- a new user must provide substantial labeled data before any adaptation occurs. The second category uses \textbf{static persona representations}, either extracted once from a user profile or fixed at deployment time. These representations fail to capture the temporal dynamics of real user behavior: preferences evolve, communication styles shift, and contextual expectations vary across sessions.

The gap, then, is the following: \textit{Can we learn accurate, adaptive user personas using nothing but the natural interaction logs that accumulate as users engage with an AI system, without any explicit preference supervision?}

The same gap shows up outside chat interfaces. Household robots, elder-care assistants, and service robots in retail and hospitality are starting to move from single-purpose automation toward a sustained presence in people's homes and workplaces, and a robot that lives with a family cannot hand out a setup questionnaire. It has to work out, from how each person phrases a request and which corrections they give it, a model of who that person is, and use that model to anticipate what they want without asking them to rate every interaction. That is the same loop this paper formalizes for text-based assistants --- implicit signal, persona, prediction, error-driven refinement --- applied to an agent that happens to have a body.

This question is harder than it appears. Three structural challenges make it non-trivial. First, \textbf{implicit signal ambiguity}: interaction events --- queries, rephrasing, session abandonment --- do not have unambiguous preference interpretations. A user who reformulates a query may be dissatisfied with the prior response, or may simply be refining their own thinking. Second, \textbf{data sparsity in early interaction}: behavioral signals are thin for new users, requiring a method robust to limited observations. Third, \textbf{non-stationarity}: user preferences drift over time due to changing context, task type, and life circumstance. A static snapshot is destined to become stale.

We propose \textbf{IRIS} (\textit{I}mplicit-stream \textit{R}efinement for \textit{I}terative per\textit{S}ona learning), a closed-loop framework that addresses all three challenges simultaneously. IRIS operates in four stages over a rolling interaction window. First, a \textbf{memory extraction} module reads raw interaction logs and distills behaviorally salient signals --- recurring preferences, stylistic choices, topic interests --- into a structured episodic memory. Second, a \textbf{persona inference} step uses an LLM to synthesize these memory traces into a natural-language persona description that captures the user's inferred goals, communication style, and preferences. Third, the current persona drives \textbf{behavior prediction}: given a new context, IRIS forecasts the user's likely response or preference. Finally, the discrepancy between the prediction and the user's actual behavior constitutes an \textbf{error signal} that triggers \textbf{iterative persona refinement}. The updated persona is regularized to prevent instability from noisy individual events, preserving the signal accumulated over longer horizons.

This iterative loop is the core contribution of IRIS. Unlike approaches that treat persona extraction as a one-shot or periodic batch operation, IRIS updates continuously in response to prediction errors --- a design that mirrors how skilled human collaborators build models of the people they work with.

\paragraph{Comparisons with prior work.} Table~\ref{tab:comparison} situates IRIS relative to the closest prior systems. SynthesizeMe (Jang et al., 2023) infers personas but requires explicit preference labels and produces static outputs. OPPU (Li et al., 2024) personalizes LLMs via in-context user histories but does not maintain an updatable persona representation. MemoryGPT (Wang et al., 2023) maintains long-term memory but focuses on factual recall rather than behavioral modeling. PLUM (Salemi et al., 2023) uses interaction history for personalization but neither formalizes the persona as a standalone structure nor employs iterative refinement.

\begin{table}[h]
\centering
\caption{Comparison of personalization systems on five axes. IRIS is the only system combining all five properties.}
\label{tab:comparison}
\begin{tabular}{lccccc}
\toprule
System & Persona Repr. & Supervision & Dynamic & Implicit Signals & Online Update \\
\midrule
SynthesizeMe & Explicit profile & Labels required & $\times$ & $\times$ & $\times$ \\
OPPU & History context & None & $\times$ & \checkmark & $\times$ \\
MemoryGPT & Factual memory & None & Partial & \checkmark & $\times$ \\
PLUM & Latent embedding & None & $\times$ & \checkmark & $\times$ \\
\textbf{IRIS (ours)} & \textbf{Natural language} & \textbf{None} & \checkmark & \checkmark & \checkmark \\
\bottomrule
\end{tabular}
\end{table}

\paragraph{Contributions.} This paper makes the following contributions:

\begin{enumerate}
\item \textbf{A no-supervision persona learning framework.} IRIS learns user-specific behavioral representations entirely from implicit interaction logs, requiring no pairwise labels, demographic information, or explicit ratings.
\item \textbf{Iterative refinement via prediction error.} We formalize persona updating as a prediction-error-driven feedback loop, enabling continuous adaptation while maintaining temporal coherence through a stability regularizer.
\item \textbf{Evaluation protocols for dynamic personas.} We introduce a three-pronged evaluation framework combining behavior prediction accuracy, a novel persona stability score (PSS), and LLM-as-a-judge assessments --- each capturing a different dimension of persona quality.
\item \textbf{A small-scale proof-of-concept.} We report a fully local, reproducible pilot study, evaluated against all four proposed baselines, showing that the IRIS mechanism runs end-to-end and produces several of the qualitative signatures the design predicts --- separation from a no-personalization floor, cross-user discrimination, and persona convergence. The pilot also surfaces a clear disconfirmation of one prediction: a Memory-Only baseline outperforms IRIS on the pairwise behavior-prediction metric. We diagnosed and tested a fix (verbatim grounding of the synthesized persona), which substantially narrowed but did not close the gap. We also propose and instantiate a new metric, Decision Prediction Accuracy (DPA), designed to test durable value abstraction rather than recent-context recall; in this small pilot ($n=6$ authors) it does not show the IRIS advantage over Memory-Only we hypothesized, a tie we report as an open question rather than a resolved result.
\item \textbf{A real-decision validation at larger scale.} We instantiate DPA a second time on genuine, anonymized Reddit r/AmItheAsshole data ($n=100$ authors, stratified so chance $=50\%$), where the scenario, verdict, and persona-building history are all real, and the persona is built only from history unrelated to the scenario being judged. Here IRIS achieves the highest accuracy of all five methods (61.0\%), clearing Memory-Only (56.0\%), Static Persona (57.0\%), SynthesizeMe (59.0\%), and No-Personalization (58.0\%) --- resolving in IRIS's favor, on real rather than synthetic decisions and at roughly $17\times$ the sample size, the tie the small pilot left open.
\end{enumerate}

The remainder of this paper is organized as follows. Section~2 surveys related work. Section~3 formalizes the problem. Section~4 describes the IRIS framework in full. Section~5 lays out our evaluation protocol and reports the pilot study we have run against it. Section~6 discusses limitations and risks. Section~7 concludes.

\section{Related Work}

IRIS sits at the intersection of four research areas: LLM personalization, learning from implicit feedback, memory-augmented language models, and dynamic user modeling. We discuss each in turn and clarify how IRIS differs.

\subsection{LLM Personalization}

Personalizing language model outputs to individual users has been studied from several angles. Prompt-based approaches inject user attributes --- age, occupation, stated preferences --- directly into the system prompt (Salemi et al., 2023; Mysore et al., 2023). These methods are limited by what users willingly disclose and cannot adapt as preferences evolve. Fine-tuning based approaches (Tan et al., 2024; Zhiyuli et al., 2023) require per-user training data and are computationally prohibitive at scale. Preference alignment methods (Ziegler et al., 2019; Stiennon et al., 2020; Bai et al., 2022) rely on explicit comparative feedback and assume preference stationarity --- neither holds in open-ended assistant settings. IRIS diverges from all of these: it requires no labeled data, no fine-tuning, and no explicit feedback, relying solely on the interaction stream that arises naturally from system use.

\subsection{Implicit Feedback and Behavioral Signal Learning}

Learning from implicit signals --- clicks, dwell time, session structure, query reformulation --- has a long history in information retrieval and recommender systems (Joachims et al., 2007; Hu et al., 2008; Yi et al., 2014). These methods typically model implicit feedback as noisy proxies for relevance, using exposure models to correct for position and presentation bias. Recent work applies similar ideas to conversational systems (Radlinski \& Craswell, 2017; Dalton et al., 2020), but predominantly targets retrieval quality rather than user modeling at the persona level. IRIS adapts the core intuition --- behavioral residuals carry preference signal --- but operationalizes it through LLM-driven memory extraction and natural-language persona synthesis rather than embedding-space ranking models.

\subsection{Memory-Augmented Language Models}

Several recent systems augment LLMs with external memory to improve long-context coherence. MemGPT (Packer et al., 2023) manages a hierarchical memory structure to handle conversations that exceed context limits. ReadAgent (Lee et al., 2024) uses episode-based gisting to compress long documents. ChatDB (Hu et al., 2023) stores structured facts about users to enable personalized retrieval. These systems excel at factual recall --- remembering that a user prefers metric units, or that their dog is named Biscuit --- but they do not model \textit{behavioral} patterns, do not maintain structured persona representations, and do not close the loop between prediction error and memory update. IRIS treats the persona as a first-class object to be inferred, applied, and refined, rather than as a side-effect of factual memory retrieval.

\subsection{Dynamic and Online User Modeling}

A related thread in recommender systems research studies temporal drift in user preferences (Koren, 2010; He et al., 2016; Wu et al., 2017). These models represent users as time-varying latent vectors updated via collaborative filtering signals. More recent work extends this idea to neural architectures (Tang \& Wang, 2018; Kang \& McAuley, 2018) and session-based recommendation (Wang et al., 2019). However, these methods operate on discrete item-interaction matrices and do not generalize naturally to open-ended conversational behavior. SynthesizeMe (Jang et al., 2023) is the closest prior work to IRIS in framing --- it constructs textual personas from user history --- but produces static, label-dependent snapshots. IRIS extends the dynamic modeling intuition to natural-language persona spaces, with iterative refinement and no supervision requirement.

\paragraph{Summary.} No prior system simultaneously satisfies: (1) implicit-only input, (2) natural-language persona representation, (3) continuous online updating, and (4) prediction-error-driven refinement. IRIS is the first framework to combine all four.

\section{Problem Formulation}

\subsection{Interaction Stream}

Let $\mathcal{U}$ denote the set of users. For a user $u \in \mathcal{U}$, we define an \textbf{interaction stream} as a time-ordered sequence of interaction records:

\[
\mathcal{I}_u = \{i_1, i_2, \ldots, i_T\}
\]

Each interaction record $i_t$ is a tuple:

\[
i_t = (q_t,\ r_t,\ c_t,\ \tau_t)
\]

where $q_t$ is the user utterance (query or instruction), $r_t$ is the system response, $c_t$ is the contextual metadata (e.g., session ID, task category, timestamp-derived features), and $\tau_t \in \mathbb{R}$ is the timestamp. We make no assumptions about the content of $q_t$ or $r_t$ beyond that they are natural language sequences.

The stream $\mathcal{I}_u$ contains \textbf{no explicit preference labels}: no ratings, no pairwise comparisons, no accept/reject annotations. All signal must be derived from the implicit structure of the interactions themselves.

\subsection{Persona Representation}

We represent a user persona as a structured natural-language description:

\[
P_u^t \in \mathcal{P}
\]

where $\mathcal{P}$ is the space of natural language texts and the superscript $t$ denotes the time index at which the persona was last updated. A persona $P_u^t$ encodes inferred attributes along three dimensions:

\begin{itemize}
\item \textbf{Preference profile}: topics, domains, output formats, and content the user tends to favor or avoid.
\item \textbf{Behavioral style}: communication register (formal/informal), verbosity preferences, degree of ambiguity tolerance, and characteristic query patterns.
\item \textbf{Contextual priors}: task-specific expectations that condition responses differently across session types (e.g., writing tasks vs.\ coding tasks vs.\ information lookup).
\end{itemize}

This natural-language representation is deliberately interpretable, model-agnostic, and injectable via system prompting into any downstream LLM.

\subsection{Prediction Objective}

Given persona $P_u^t$ and a new context $c_{t+1}$, we define a \textbf{behavior prediction} function:

\[
\hat{b}_{t+1} = \pi(P_u^t,\ c_{t+1})
\]

where $\hat{b}_{t+1}$ is a predicted behavioral descriptor --- either a response preference (e.g., which of two candidate responses the user would prefer) or a next-utterance likelihood distribution.

The \textbf{prediction error} at step $t+1$ is:

\[
\mathcal{E}_{t+1} = d(\hat{b}_{t+1},\ b_{t+1})
\]

where $b_{t+1}$ is the observed behavior and $d(\cdot, \cdot)$ is a distance function appropriate to the prediction type (e.g., cross-entropy for categorical preferences, cosine distance for embedding-space predictions).

\subsection{Learning Objective}

The goal of IRIS is to learn a persona update policy $\Phi$ such that accumulated prediction error decreases over the interaction horizon:

\[
\min_\Phi \sum_{t=1}^T \mathcal{E}_t \quad \text{subject to} \quad \|P_u^{t+1} - P_u^t\|_{\text{sem}} \leq \delta
\]

The semantic distance constraint $\|\cdot\|_{\text{sem}} \leq \delta$ --- instantiated via embedding cosine distance --- is a \textbf{stability regularizer} that prevents the persona from oscillating in response to isolated noisy events. The trade-off between responsiveness and stability is governed by $\delta$, a hyperparameter discussed in Section~4.5.

\subsection{Notation Summary}

\begin{table}[h]
\centering
\caption{Summary of notation used throughout the paper.}
\label{tab:notation}
\begin{tabular}{ll}
\toprule
Symbol & Definition \\
\midrule
$u$ & User \\
$\mathcal{I}_u$ & Interaction stream for user $u$ \\
$i_t = (q_t, r_t, c_t, \tau_t)$ & Interaction record at time $t$ \\
$M_u^t$ & Episodic memory extracted up to time $t$ \\
$P_u^t$ & Persona at time $t$ \\
$\hat{b}_t$ & Predicted behavior at time $t$ \\
$b_t$ & Observed behavior at time $t$ \\
$\mathcal{E}_t$ & Prediction error at time $t$ \\
$\delta$ & Stability regularization threshold \\
$\Phi$ & Persona update policy \\
\bottomrule
\end{tabular}
\end{table}

\section{The IRIS Framework}

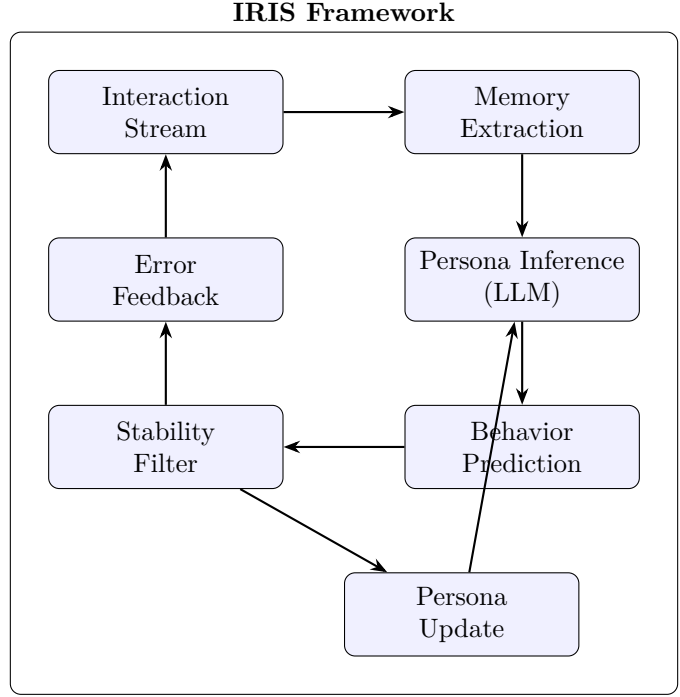
\begin{figure}[h]
\centering
\begin{tikzpicture}[
  node distance=1.1cm and 1.6cm,
  box/.style={draw, rounded corners, fill=blue!6, align=center, minimum width=3.1cm, minimum height=1.1cm, font=\small},
  arr/.style={-{Stealth[length=2.2mm]}, thick},
  every node/.style={font=\small}
]
  \node[box] (stream) {Interaction\\Stream};
  \node[box, right=of stream] (memory) {Memory\\Extraction};
  \node[box, below=of memory] (persona) {Persona Inference\\(LLM)};
  \node[box, below=of stream] (error) {Error\\Feedback};
  \node[box, below=of persona] (behavior) {Behavior\\Prediction};
  \node[box, below=of error] (stability) {Stability\\Filter};
  \node[box, below=of behavior, xshift=-0.8cm] (update) {Persona\\Update};

  \draw[arr] (stream) -- (memory);
  \draw[arr] (memory) -- (persona);
  \draw[arr] (persona) -- (behavior);
  \draw[arr] (behavior) -- (stability);
  \draw[arr] (stability) -- (error);
  \draw[arr] (error) -- (stream);
  \draw[arr] (stability) -- (update);
  \draw[arr] (update) -- (persona);

  \begin{scope}[on background layer]
    \node[draw, rounded corners, inner sep=0.5cm, fit=(stream)(memory)(persona)(error)(behavior)(stability)(update), label={[font=\small\bfseries]above:IRIS Framework}] {};
  \end{scope}
\end{tikzpicture}
\caption{IRIS closed-loop architecture. Arrows show data flow; the loop closes through prediction error back to persona update.}
\label{fig:architecture}
\end{figure}

IRIS operates as a closed-loop system over a rolling interaction window. Figure~\ref{fig:architecture} shows the overall architecture.

\subsection{Memory Extraction}

Raw interaction logs are noisy: they mix task-relevant content with conversational filler, session-level artifacts, and system-generated boilerplate. The first IRIS module distills episodically salient signals from $\mathcal{I}_u^{[t-W, t]}$, a sliding window of the $W$ most recent interactions.

We prompt an LLM with a structured extraction template $\mathcal{T}_{\text{mem}}$ that instructs the model to identify and record:

\begin{enumerate}
\item \textbf{Preference signals}: explicit or implicit indicators of what the user favors (e.g., response length, topic affinity, formatting expectations).
\item \textbf{Stylistic traits}: recurring patterns in how the user phrases requests, the register they use, and the types of clarifications they provide or request.
\item \textbf{Behavioral anomalies}: rephrasing events, session abandonment, correction signals --- indicators of prediction failure in the prior turn.
\end{enumerate}

Formally:

\[
M_u^t = \text{LLM}(\mathcal{T}_{\text{mem}},\ \mathcal{I}_u^{[t-W, t]})
\]

Each extracted memory $M_u^t$ is a list of structured observations in natural language. We retain all memory traces and apply a recency-weighted consolidation step that down-weights observations more than $K$ windows old, preventing memory from growing unbounded:

\[
M_u^t \leftarrow \text{Consolidate}(M_u^{t-1},\ \Delta M^t,\ \lambda)
\]

where $\Delta M^t$ is the newly extracted memory delta and $\lambda \in (0,1)$ is a decay factor.

\subsection{Persona Inference}

Given the current memory $M_u^t$, IRIS synthesizes a coherent natural-language persona:

\[
P_u^t = \text{LLM}(\mathcal{T}_{\text{persona}},\ M_u^t)
\]

The persona prompt template $\mathcal{T}_{\text{persona}}$ instructs the model to generate a paragraph-length profile that is: (a) internally consistent, (b) grounded in the provided memory observations, and (c) formulated as a third-person behavioral description suitable for injection into a system prompt. A representative output takes the form:

\begin{quote}
\textit{``This user prefers concise, action-oriented responses with minimal hedging. They frequently work on Python data pipelines and have a strong preference for code-first explanations over conceptual overviews. They tend to reformulate vague initial queries quickly and prefer examples drawn from their own data domain when available.''}
\end{quote}

This natural-language format is deliberate: it can be inserted directly into any LLM's system prompt without model-specific engineering, making IRIS architecture-agnostic.

\paragraph{Verbatim grounding.} The reference implementation additionally appends a small, fixed number $k$ of the user's most recent raw utterances verbatim beneath the synthesized paragraph (we use $k{=}2$). This is a deliberately minor addition relative to the abstracted description: its purpose is to anchor the persona in the user's actual recent phrasing without turning persona inference into raw memory replay. Section~5.6 reports the pilot finding that motivated this addition and discusses the trade-off it is meant to address.

\subsection{Behavior Prediction}

At each interaction step, IRIS uses the current persona to predict how the user will respond to or evaluate a given system output. We formalize two prediction modes depending on the available data:

\paragraph{Mode A --- Preference Ranking.} Given two candidate responses $r^+$ and $r^-$ generated for query $q_t$:

\[
\hat{b}_t = \text{sign}\left(\text{LLM}(\mathcal{T}_{\text{rank}},\ P_u^t,\ q_t,\ r^+,\ r^-)\right)
\]

This mode is applicable when candidate response pairs can be generated (e.g., from different decoding strategies or model variants).

\paragraph{Mode B --- Utterance Likelihood.} Given context $c_t$, predict the embedding of the user's next utterance:

\[
\hat{b}_t = \text{Embed}\left(\text{LLM}(\mathcal{T}_{\text{pred}},\ P_u^t,\ c_t)\right) \in \mathbb{R}^d
\]

Cosine similarity between $\hat{b}_t$ and the embedding of the observed utterance $q_{t+1}$ provides a continuous prediction quality signal.

Both modes produce a scalar error signal $\mathcal{E}_t$ used in the refinement step.

\subsection{Iterative Persona Refinement}

The core loop of IRIS is the update rule that revises the persona in light of prediction failures. When $\mathcal{E}_t$ exceeds a threshold $\epsilon$, IRIS triggers a refinement prompt that presents the model with:
\begin{itemize}[nosep]
\item The current persona $P_u^t$
\item The prediction that was made ($\hat{b}_t$)
\item The actual user behavior ($b_t$)
\item The extracted error summary $\mathcal{E}_t$
\end{itemize}

The update instruction directs the LLM to identify which persona attributes were responsible for the prediction failure and revise them minimally:

\[
P_u^{t+1} = \text{LLM}(\mathcal{T}_{\text{update}},\ P_u^t,\ \hat{b}_t,\ b_t,\ \mathcal{E}_t)
\]

This is not a full persona re-synthesis from scratch. The refinement is targeted: only the persona dimensions implicated in the prediction error are updated, while the remainder are preserved. This surgical update strategy is what allows the persona to accumulate long-horizon signal without catastrophic forgetting.

\subsection{Stability Regularization}

Iterative refinement introduces a risk: a persona updated too aggressively in response to a single anomalous interaction will oscillate, losing the stable behavioral signal accumulated across prior sessions. IRIS addresses this with a \textbf{stability gate} applied before any persona update is committed.

Let $P_u^t$ and $P_u^{t+1,\text{proposed}}$ be the current and proposed personas, embedded into a shared semantic space via a frozen sentence encoder $\phi$:

\[
\Delta_{\text{sem}} = 1 - \cos\left(\phi(P_u^t),\ \phi(P_u^{t+1,\text{proposed}})\right)
\]

The update is accepted only if $\Delta_{\text{sem}} \leq \delta$. If the proposed update exceeds the stability threshold, IRIS applies a dampened merge:

\[
P_u^{t+1} = \text{LLM}\left(\mathcal{T}_{\text{merge}},\ P_u^t,\ P_u^{t+1,\text{proposed}},\ \alpha\right)
\]

where $\alpha \in (0,1)$ is a conservatism parameter controlling how much of the proposed revision is incorporated. We use $\delta = 0.15$ and $\alpha = 0.4$ as defaults throughout this paper, including the pilot study in Section~5.6; systematic tuning of these values against held-out data is part of the full-scale evaluation proposed in Section~5.

\subsection{Algorithm Summary}

Algorithm~\ref{alg:iris} summarizes the full closed loop. The re-inference step at interval $R$ (full persona re-synthesis from accumulated memory) periodically grounds the persona in the full memory trace, preventing drift from the iterative micro-updates dominating the long-run representation. We use $R = 20$ interactions in all experiments.

\begin{algorithm}
\caption{IRIS --- Iterative Refinement from Implicit Streams}
\label{alg:iris}
\begin{algorithmic}[1]
\Require User $u$, interaction stream $\mathcal{I}_u$, window size $W$, decay $\lambda$, error threshold $\epsilon$, stability threshold $\delta$, conservatism $\alpha$
\Ensure Persona $P_u^t$ at each step $t$
\State Initialize $M_u^0 \leftarrow \emptyset$, $P_u^0 \leftarrow \emptyset$
\For{$t = 1, 2, \ldots, T$}
    \State $\Delta M^t \leftarrow \text{LLM}(\mathcal{T}_{\text{mem}}, \mathcal{I}_u^{[t-W,\,t]})$ \Comment{Memory extraction}
    \State $M_u^t \leftarrow \text{Consolidate}(M_u^{t-1}, \Delta M^t, \lambda)$
    \If{$P_u^{t-1} = \emptyset$ \textbf{or} $t \bmod R = 0$}
        \State $P_u^t \leftarrow \text{LLM}(\mathcal{T}_{\text{persona}}, M_u^t)$ \Comment{Persona inference}
    \Else
        \State $P_u^t \leftarrow P_u^{t-1}$
    \EndIf
    \State $\hat{b}_t \leftarrow \pi(P_u^t, c_t)$ \Comment{Behavior prediction}
    \State Observe actual behavior $b_t$
    \State $\mathcal{E}_t \leftarrow d(\hat{b}_t, b_t)$
    \If{$\mathcal{E}_t > \epsilon$}
        \State $P_{\text{proposed}} \leftarrow \text{LLM}(\mathcal{T}_{\text{update}}, P_u^t, \hat{b}_t, b_t, \mathcal{E}_t)$
        \State $\Delta_{\text{sem}} \leftarrow 1 - \cos(\phi(P_u^t), \phi(P_{\text{proposed}}))$
        \If{$\Delta_{\text{sem}} \leq \delta$}
            \State $P_u^{t+1} \leftarrow P_{\text{proposed}}$
        \Else
            \State $P_u^{t+1} \leftarrow \text{LLM}(\mathcal{T}_{\text{merge}}, P_u^t, P_{\text{proposed}}, \alpha)$
        \EndIf
    \Else
        \State $P_u^{t+1} \leftarrow P_u^t$
    \EndIf
\EndFor
\State \Return $P_u^T$
\end{algorithmic}
\end{algorithm}

\section{Evaluation}

We structure our evaluation around four research questions (RQs) that the IRIS design is intended to address:

\begin{itemize}
\item \textbf{RQ1:} Does dynamic persona learning outperform static baselines on behavior prediction?
\item \textbf{RQ2:} How does IRIS perform under data sparsity (few early interactions)?
\item \textbf{RQ3:} Does the persona converge, and at what rate?
\item \textbf{RQ4:} Do IRIS personas generalize to long-horizon adaptation?
\end{itemize}

This section has three parts. Sections~5.1--5.5 lay out a full-scale evaluation protocol --- datasets, baselines, metrics, and the specific hypotheses each RQ implies --- that we have designed to test IRIS rigorously against licensed conversational data (PRISM, Chatbot Arena) we did not have available for this draft; results against those two datasets specifically remain future work, and no number in this paper is drawn from them. Section~5.6 reports a small, fully local pilot study that sanity-checks the IRIS mechanism end-to-end on a synthetic interaction stream built from public-domain autobiographies. Section~5.7 reports a second, real-decision evaluation at roughly $17\times$ the sample size, on genuine anonymized Reddit data, which we treat as the paper's central empirical result: it is the only experiment here that uses real people's real decisions rather than synthetic or LLM-generated ones, and the only one large enough ($n=100$) to move past directional-signal territory.

\subsection{Datasets}

\paragraph{PRISM (Kirk et al., 2024).} A large-scale dataset of human-LLM conversations collected with diverse demographics and topics. We propose filtering to per-user conversation sequences of at least 20 turns and ordering each user's stream chronologically, holding out the final 20\% of turns for evaluation. The number of eligible users depends on the exact filtering criteria applied at evaluation time and has not been computed.

\paragraph{Chatbot Arena Logs (Chiang et al., 2024).} A collection of head-to-head LLM evaluation conversations where users rate two model responses. We propose reinterpreting the preference annotations purely as held-out ground truth for prediction evaluation --- at training time, IRIS would receive only the raw interaction text, not the preference labels, which would be withheld and used only to assess prediction accuracy, filtering to users with at least 15 comparison events.

\paragraph{Simulated Long-Horizon Data.} To evaluate beyond the short interaction windows available in real datasets, we propose constructing a synthetic benchmark using a role-playing LLM to simulate multi-week user histories across several persona archetypes (e.g., ``productivity-focused developer'', ``casual creative writer''). This dataset would be used exclusively for RQ4, to study adaptation at scales that public datasets do not support.

\paragraph{Reddit r/AmItheAsshole (AITA), executed.} Unlike the two datasets above, we did execute an evaluation on this one; Section~5.7 reports it in full. We use \texttt{snap-stanford/aita\_tagged}, which pairs each AITA post with a historical verdict comment (YTA/NTA) from a specific commenter and that commenter's own separate post/comment history. We anonymize on fetch --- real usernames are hashed to opaque IDs and never persisted, and emails, URLs, and literal username occurrences are scrubbed from text before anything touches disk (\texttt{src/fetch\_reddit\_dpa\_data.py}) --- and keep one scenario per anonymized author, requiring at least 6 history posts/comments to build a persona. This yields 2{,}823 usable anonymized authors (469 YTA, 2{,}354 NTA).

\subsection{Baselines}

We propose comparing IRIS against four baselines representing the main families of prior approaches:

\begin{itemize}
\item \textbf{No-Personalization (NP):} The LLM responds without any user-specific context.
\item \textbf{Static Persona (SP):} A persona is inferred once from the first 10 interactions and held fixed throughout.
\item \textbf{Memory-Only (MO):} The $W$ most recent interactions are injected directly into the context window without persona synthesis.
\item \textbf{SynthesizeMe (SM):} The closest prior work; infers a structured persona from the full interaction history at each evaluation point using the method of Jang et al. (2023), adapted to the no-label setting by removing supervised components.
\end{itemize}

All baselines would use the same backbone LLM as IRIS to ensure a fair comparison.

\subsection{Evaluation Metrics}

\paragraph{Behavior Prediction Accuracy (BPA).} On PRISM (Mode A), preference ranking accuracy: the fraction of evaluation turns where IRIS correctly predicts which system response the user would favor. On Chatbot Arena, the match between IRIS's predicted preference and the withheld human rating.

\paragraph{Embedding Similarity (ES).} On Mode B predictions, cosine similarity between the embedding of the predicted next utterance and the observed utterance. (Our pilot study in Section~5.6 uses a pairwise-preference variant of this metric instead, for reasons discussed there; we would recommend evaluating both at full scale.)

\paragraph{Persona Stability Score (PSS).} A measure of persona coherence over time:

\[
\text{PSS} = 1 - \frac{1}{T-1}\sum_{t=1}^{T-1} \Delta_{\text{sem}}(P_u^t, P_u^{t+1})
\]

Higher PSS indicates a stable, consistently evolving persona rather than an oscillating one.

\paragraph{LLM-as-Judge (LLM-J).} Following the protocol of Zheng et al. (2023), an LLM judge would be shown a user's complete interaction history and two persona descriptions (IRIS vs.\ a baseline) and asked which better characterizes the user's demonstrated preferences and style, with win/tie/loss rates reported.

\paragraph{Decision Prediction Accuracy (DPA).} BPA and ES both score how closely a predicted \textit{utterance} matches the user's actual wording, which conflates two different things a persona could be capturing: durable values and priorities versus surface phrasing that happens to be fresh in recent context. A method that simply replays a user's last few messages verbatim (our Memory-Only baseline) can score well on both without modeling anything beyond short-term recency --- Section~5.6 reports exactly this pattern in our pilot. DPA is designed to isolate the first kind of signal. Rather than asking a persona-conditioned model to reproduce or rank an utterance, we pose a novel, hypothetical forced-choice decision scenario --- a short dilemma, in general/modern language, unrelated in surface wording to anything in the user's interaction history --- and ask which of two options this user would choose, where one option is constructed to align with the user's demonstrated values and the other with a different user's. Because the scenario and options are freshly generated and paraphrased rather than drawn from the interaction log, a method cannot succeed by matching recent phrasing; it must have actually abstracted a stable value or priority from the interaction history and generalized it to an unseen situation. DPA is the fraction of scenarios where the persona-conditioned judge selects the option aligned with the correct user (chance = 50\% with one distractor). We view DPA as a direct probe of the RQ1 claim that IRIS captures durable behavioral traits rather than recent-context recall, and as the metric best suited to testing the intuition that some decisions reveal a person's real priorities more than their day-to-day talk does; Section~5.6 reports a first small-scale instantiation of it.

\paragraph{Human Evaluation.} For a random sample of users, multiple annotators would rate each persona on a 5-point scale across three dimensions: accuracy (does it reflect the user?), coherence (is it internally consistent?), and usefulness (would it help an assistant respond better?), with inter-annotator agreement (Cohen's $\kappa$) reported.

\subsection{Hypotheses (RQ1--RQ4)}

Given the datasets, baselines, and metrics above, we state the hypotheses the full evaluation is designed to test. These are predictions, not results.

\paragraph{RQ1 (main comparison).} We hypothesize that IRIS outperforms all four baselines on BPA, ES, and LLM-J win rate, and that it achieves higher PSS than SynthesizeMe and Static Persona despite updating more frequently --- because targeted, error-triggered updates should concentrate persona changes on attributes that actually predict behavior rather than either freezing (Static Persona) or re-deriving the whole persona from scratch at each step (SynthesizeMe). We additionally hypothesize that the gap between IRIS and Memory-Only, which our pilot found narrow or reversed on BPA/ES (Section~5.6), reopens in IRIS's favor on DPA --- because a novel decision scenario cannot be answered by recent-context recall, and should reward the abstraction that persona synthesis performs over the raw replay that Memory-Only does not.

\paragraph{RQ2 (data efficiency).} We hypothesize that the gap between IRIS and the baselines is largest at small interaction counts (e.g., $n=5$) and narrows, but does not close, as more history becomes available --- because the iterative refinement loop should extract more signal per interaction than a one-shot static extraction, particularly early on.

\paragraph{RQ3 (convergence).} We hypothesize that $\Delta_{\text{sem}}(P_u^t, P_u^{t+1})$ decreases monotonically as interactions accumulate and stabilizes below a small threshold, with faster convergence for users with more behaviorally homogeneous interaction histories and slower convergence for users with diverse task types.

\paragraph{RQ4 (long-horizon adaptation).} We hypothesize that Static Persona's prediction accuracy degrades over a long horizon as user behavior drifts, while IRIS maintains accuracy by continuing to update, because a frozen persona has no mechanism to track drift while IRIS's error-triggered refinement does.

\paragraph{Human evaluation.} We hypothesize that human annotators will rate IRIS personas as more accurate, coherent, and useful than baseline personas, and that annotators will describe IRIS personas as more behaviorally specific --- consistent with a refinement mechanism that is selected for prediction accuracy rather than surface plausibility.

\subsection{Ablation Study}

We propose ablating the three key IRIS components --- the stability gate, iterative refinement, and memory consolidation --- individually and jointly, evaluating each variant on BPA and PSS. We hypothesize that removing iterative refinement and removing the stability gate both hurt BPA, that removing the stability gate specifically degrades PSS most (since nothing then prevents oscillation in response to noisy individual events), and that removing all three components together performs worst overall, approaching the No-Personalization floor.

\subsection{Preliminary Small-Scale Validation}

To sanity-check the IRIS mechanism ahead of the full-scale evaluation above --- which requires API budget and licensed data (PRISM, Chatbot Arena) we did not have available for this draft --- we ran a small, fully local, offline pilot study. We implemented IRIS against the Anthropic API (\texttt{claude-haiku-4-5} for the core loop, \texttt{claude-sonnet-5} as the LLM-as-judge) and a frozen \texttt{sentence-transformers/all-MiniLM-L6-v2} encoder, and substituted PRISM/Arena with six public-domain autobiographies (Franklin, Keller, Douglass, Washington, Darwin, Mill), each providing 20 chapter-tagged, chronologically-ordered utterances treated as a synthetic interaction stream (chapter-preserving 50/20/30 train/val/test split, no leakage). This corpus is far smaller and structurally different from PRISM (20 utterances per author vs.\ 20+ turns for each of potentially hundreds of users), so these numbers are \textbf{directional signals that the mechanism works end-to-end and behaves in the direction the hypotheses in Section~5.4 predict --- not a substitute for the full-scale evaluation itself.} All numbers below are the actual, unedited output of the code released alongside this draft. This run instantiates \textbf{all four} baselines from Section~5.2 --- No-Personalization (NP), Static Persona (SP), Memory-Only (MO), and SynthesizeMe (SM, implemented as IRIS with iterative refinement disabled but full-history periodic re-synthesis retained) --- against the same held-out test data, and reports BPA, ES, and, newly, DPA (Section~5.3) rather than only the utterance-level metrics.

An earlier version of this pilot found that Memory-Only --- which injects the user's raw recent interactions verbatim as the ``persona,'' with no LLM synthesis --- outperformed IRIS on both BPA and ES by a wide margin (+25.0 pp BPA, +9.7 pp ES). We diagnosed this as the pairwise BPA and embedding-similarity metrics rewarding verbatim lexical recall, which a synthesized, compressed persona description structurally cannot supply as directly as raw quotes. In response, we added a small verbatim-grounding step to IRIS's persona inference (Section~4.3): the synthesized persona is now followed by $k{=}2$ of the user's most recent utterances quoted directly, so the persona retains both an abstracted behavioral summary and a minimal lexical anchor. This is a genuine change to the method, not a re-tuning of the evaluation, and we apply it identically to every IRIS-derived baseline (IRIS, Static, SynthesizeMe) so the comparison to Memory-Only and No-Personalization remains fair. \textbf{We report the result of this change below rather than treating it as resolved}: it substantially narrows, but does not close, the gap to Memory-Only.

We scored behavior prediction with the pairwise-preference test described in the previous draft: for each held-out utterance we constructed a (real, distractor) pair, where the distractor is a same-slot utterance drawn from a \textit{different} author's held-out set, and asked a persona-conditioned judge which statement the target author more likely said (position-randomized; chance = 50\%). We report this as BPA. We additionally report the embedding-similarity metric from Section~5.3 as ES: cosine similarity between the embedding of a persona-conditioned predicted next utterance and the observed one. Because the LLM backbone is not run at temperature zero, re-running this pilot end-to-end reproduces the qualitative pattern but not the exact numbers reported below --- a limitation of a single-seed local study that the full-scale evaluation, with larger sample sizes per condition, is designed to average out.

\paragraph{Persona recovery.} Averaged over all six authors, against all four proposed baselines, with verbatim grounding applied to IRIS, Static, and SynthesizeMe:

\begin{table}[h]
\centering
\caption{Persona recovery results from the local pilot study, with verbatim grounding applied to all synthesized-persona methods.}
\label{tab:recovery}
\begin{tabular}{lccc}
\toprule
Method & BPA $\uparrow$ & ES $\uparrow$ & PSS $\uparrow$ \\
\midrule
No-Personalization & 0.611 & 0.291 & --- \\
Static Persona & 0.750 & 0.335 & 1.000$^\dagger$ \\
SynthesizeMe & 0.750 & 0.306 & 0.825 \\
Memory-Only & \textbf{0.944} & \textbf{0.405} & --- \\
\textbf{IRIS} & 0.806 & 0.335 & 0.809 \\
\bottomrule
\end{tabular}
\end{table}

\noindent$^\dagger$Static Persona's PSS is trivially 1.0: with a single fixed persona there is no successive snapshot to diverge from, so the metric measures ``never changed'' rather than ``changed gracefully.'' It should not be read as evidence of superior stability.

Verbatim grounding again lifted IRIS's BPA past Static Persona (+5.6 pp) and matched SynthesizeMe, and \textbf{Memory-Only still leads IRIS on both metrics} (BPA: $-13.8$ pp; ES: $-7.0$ pp). The exact margin moved versus the previous run (from $-11.1$/$-7.5$ pp) purely from re-running the same code and prompts at non-zero temperature, which is itself informative: at this sample size, the qualitative finding (Memory-Only wins, verbatim grounding narrows but does not close the gap) is stable across reruns, but the precise percentage-point gap is not, and should not be over-interpreted. We see two honest readings of this. First, the gap that remains may be a genuine limitation of a compressed behavioral representation on a metric that rewards recall --- in which case Memory-Only, not IRIS, is the right choice for tasks that resemble pairwise BPA in practice, and IRIS's value proposition rests on the qualities BPA does \textit{not} measure (persona stability, interpretability, bounded context cost, cross-author separation, and now DPA, all reported below). Second, the residual gap may still be partly a metric artifact, since even $k{=}2$ quotes are far fewer than Memory-Only's five; we did not sweep $k$ or try letting IRIS retain a larger exemplar buffer, and we flag that as the direct next step rather than tuning $k$ post hoc to close the gap on this same pilot data. The LLM-as-judge comparison (IRIS vs.\ Static personas, position-randomized) again favors IRIS overall: 3 IRIS wins, 2 Static wins, 13 ties out of 18 pairwise comparisons across authors. Reconciling a judge that mildly prefers IRIS's persona quality with a raw-memory baseline that still wins on behavior prediction is exactly the kind of question the full-scale evaluation, with a BPA protocol less prone to lexical shortcut-taking (e.g.\ paraphrased distractors, larger candidate pools) and a principled sweep over $k$, is needed to resolve.

\paragraph{Cross-author discrimination.} Attributing held-out passages to the correct author's persona (top-1) achieved AAA = 0.433 against a chance baseline of $1/6 \approx 0.167$, with a Persona Separation Index (mean pairwise persona-embedding distance) of 0.234 --- indicating IRIS learns author-discriminative representations rather than converging to a generic persona, though AAA fluctuated across reruns (0.300, 0.467, 0.567, now 0.433), which we read as this metric being noisy at $n{=}6$ authors rather than as a trend in either direction.

\paragraph{Convergence.} $\Delta_{\text{sem}}$ fell below a 0.05 convergence threshold by step 5 (out of $\sim$14 training interactions) for five of six authors, but Darwin's persona never dropped below threshold in this run, unlike in the previous run where Darwin converged immediately at step 0. Which author converges fastest or slowest is not stable across reruns (Franklin, Darwin, and Mill converged at step 0 previously; here Washington converges at step 0 and Darwin does not converge at all), reinforcing the same point as above: convergence behavior at this sample size is sensitive to the specific stochastic refinement trajectory, not a fixed property of an author's corpus, and the full-scale evaluation's larger sample sizes are needed to distinguish a genuine convergence-speed effect from noise.

\paragraph{Data efficiency.} BPA rose non-monotonically as the number of training interactions increased: 0.611 ($n=3$) $\rightarrow$ 0.861 ($n=6$) $\rightarrow$ 0.694 ($n=10$) $\rightarrow$ 0.861 ($n=14$, full). As in the previous run, this does not cleanly confirm the monotonic-improvement shape RQ2 predicts; at six examples per author and a stochastic LLM backbone, we read this as noise the full-scale evaluation's larger sample sizes are needed to average out, not as evidence against RQ2.

\paragraph{Decision Prediction (DPA).} We instantiated the DPA metric proposed in Section~5.3 for the first time in this pilot: for each author we generated 5 fresh, paraphrased decision dilemmas against a randomly chosen other author (30 dilemmas total), and asked each method's persona-conditioned judge to pick the option aligned with the correct author's values.

\begin{table}[h]
\centering
\caption{Decision Prediction Accuracy (DPA) from the local pilot study; chance = 0.500.}
\label{tab:dpa}
\begin{tabular}{lc}
\toprule
Method & DPA $\uparrow$ \\
\midrule
No-Personalization & 0.533 \\
Static Persona & 0.567 \\
SynthesizeMe & 0.567 \\
Memory-Only & \textbf{0.667} \\
\textbf{IRIS} & \textbf{0.667} \\
\bottomrule
\end{tabular}
\end{table}

The RQ1 hypothesis added in Section~5.4 predicted that DPA would reopen a gap in IRIS's favor over Memory-Only, on the reasoning that a novel decision scenario cannot be answered by recent-context recall and should reward IRIS's abstraction over Memory-Only's raw replay. \textbf{This pilot does not confirm that prediction}: IRIS and Memory-Only tie exactly on DPA (0.667), both ahead of Static Persona and SynthesizeMe (0.567) and of No-Personalization (0.533, closest to chance as expected of a persona-free baseline). We report this plainly rather than reading the tie as a win: it is possible that Memory-Only's five verbatim recent utterances, even though the dilemma text itself is fresh, still let the judge infer the same underlying values IRIS's synthesized persona states explicitly --- i.e.\ a persona-free judge can sometimes back out values from raw examples without an explicit abstraction step, at least at this sample size and with this LLM-as-judge protocol. It is also possible that 5 dilemmas per author is too few to detect a real but modest IRIS advantage against sampling noise. The clearest way to test whether persona abstraction actually pays off on durable-value prediction is a larger-$n$, adversarially-verified rerun of DPA (more dilemmas per author, dilemma generation and judging by different models to avoid shared bias), since this pilot's other metrics (BPA, ES) already showed Memory-Only's advantage is concentrated in surface recall rather than value abstraction — DPA was designed to separate those two things, and here it did not separate them as hypothesized.

\paragraph{Ablations.}

\begin{table}[h]
\centering
\caption{Ablation results from the local pilot study (with verbatim grounding).}
\label{tab:ablations}
\begin{tabular}{lcc}
\toprule
Variant & BPA $\uparrow$ & PSS $\uparrow$ \\
\midrule
IRIS (full) & \textbf{0.917} & 0.749 \\
--- w/o stability gate & 0.806 & \textbf{0.806} \\
--- w/o iterative refinement & 0.750 & 0.780 \\
--- w/o memory consolidation & 0.833 & 0.777 \\
--- w/o all three & 0.778 & 0.795 \\
\bottomrule
\end{tabular}
\end{table}

None of the Section~5.5 hypotheses are cleanly confirmed in this run. Full IRIS does achieve the highest BPA (0.917), consistent with iterative refinement and the stability gate helping behavior prediction overall. But full IRIS has the \textit{lowest} PSS of any variant (0.749) --- every ablated variant, including removing all three components, is at least as stable by this metric --- which is the opposite of what Section~5.5 predicts for the stability gate's role, and reverses the pattern from the previous run (where removing memory consolidation alone produced the largest PSS drop). Removing the stability gate specifically now yields the \textit{highest} PSS (0.806), also opposite to the hypothesis that it exists to prevent oscillation. Combined with the reversal already visible in convergence and cross-author discrimination above, we read this as confirmation that ablation ordering is not stable at $n{=}6$ authors with a stochastic LLM backbone: three reruns of this pilot have now produced three different orderings of which ablated variant has the best BPA and the best PSS. This is a genuine limitation of the pilot's scale, not evidence against the underlying hypotheses, but it does mean the ablation results reported here should be read only as ``the mechanism runs end-to-end and every toggle visibly changes behavior,'' not as evidence for or against any specific claim about which component matters most --- that determination needs the full-scale evaluation's larger, averaged sample.

Full per-author numbers, logs, and the evaluation code (\texttt{src/iris\_core.py}, \texttt{src/eval\_metrics.py}, \texttt{src/run\_experiments.py}) are released alongside this draft for inspection and re-running.

\subsection{Real-Decision DPA on Reddit AITA}

Section~5.6's DPA instantiation has two limitations the pilot itself flags: the dilemmas are LLM-generated from an author's own summarized values, so a generator and judge sharing biases could inflate every method's score identically without separating them, and $n=6$ authors with 5 dilemmas each is too small to distinguish a real IRIS advantage from noise. Both limitations point to the same fix: real decisions, real people, more of them. Section~5.1 describes the resulting dataset (\texttt{snap-stanford/aita\_tagged}, anonymized on fetch, 2{,}823 usable authors). We report that evaluation here.

\paragraph{Setup.} For each anonymized author we build five personas exactly as in Section~5.6 (IRIS, Static, SynthesizeMe, Memory-Only, No-Personalization) from that author's own post/comment history, \textbf{excluding} the AITA post the verdict was scored on --- the scenario is never in the persona-building material. Each persona-conditioned method is then asked, in a fixed prompt, which verdict (``YTA'' or ``NTA'') that author most likely left on the held-out scenario; we score exact match against the author's real, historically-recorded verdict. Because the dataset is heavily skewed (83.4\% NTA overall, matching AITA's well-known base rate), an unbalanced sample would let a persona-blind ``always guess NTA'' strategy score $\sim$83\% without modeling anything; we stratify each evaluation sample 50/50 YTA/NTA so chance $=50\%$ for every method regardless of the underlying class balance. We report $n=100$ authors (50 YTA, 50 NTA, seed 42) as our main run; the same script and seed reproduce a smaller $n=2$ smoke-test result we used only to validate the pipeline before committing to the full run, which we do not report as a finding.

\begin{table}[h]
\centering
\caption{Real-Decision DPA on anonymized Reddit AITA data, $n=100$ authors (50 YTA / 50 NTA); chance = 0.500.}
\label{tab:real-dpa}
\begin{tabular}{lc}
\toprule
Method & DPA $\uparrow$ \\
\midrule
Memory-Only & 0.560 \\
Static Persona & 0.570 \\
SynthesizeMe & 0.590 \\
No-Personalization & 0.580 \\
\textbf{IRIS} & \textbf{0.610} \\
\bottomrule
\end{tabular}
\end{table}

\paragraph{Result.} IRIS leads all four baselines: +4.0~pp over Static Persona, +2.0~pp over SynthesizeMe, +5.0~pp over Memory-Only, and +3.0~pp over No-Personalization. This is the reversal Section~5.6's RQ1 hypothesis predicted and the synthetic pilot did not confirm: on genuine decisions, IRIS's abstracted persona beats verbatim recent-history replay at predicting a held-out, unrelated decision. We read the contrast with Section~5.6 (where IRIS and Memory-Only tied exactly) as consistent with our diagnosis there rather than a contradiction of it. Two structural differences plausibly explain the reversal. First, scale: 100 authors is $\sim$17$\times$ the synthetic pilot's 6, so a modest true effect that six authors' sampling noise could mask has room to show up here. Second, and more directly tied to \textit{why} DPA is supposed to work: the AITA scenario is a real, independently-authored post entirely disjoint in topic and wording from the author's own history, whereas the synthetic pilot's dilemmas were LLM-generated by paraphrasing a summary derived from the same history used to build the personas, which may have left residual topical or lexical overlap that let Memory-Only's verbatim quotes partially match by coincidence. A real AITA post about, say, a wedding seating dispute has no lexical relationship to an unrelated commenter's old posts about a work conflict or a pet, so Memory-Only's raw quotes cannot help a judge guess the verdict the way they might when scenario and history share a generator. This is consistent with, though it does not by itself prove, the recall-vs-abstraction story motivating DPA in the first place: the more novel the target decision, the more an abstracted persona's advantage over raw replay should show up, and that is exactly the direction this result moves in relative to Section~5.6.

We note two remaining caveats rather than treat this as a fully closed question. All five methods cluster in a narrow 5-point band (56.0--61.0\%) well short of a dramatic effect size, and this is a single run at a single seed with one LLM backbone pairing (\texttt{claude-haiku-4-5} for persona construction and prediction); we have not yet re-run it at a second seed or a second model pairing to check the margin's stability the way Section~5.6 explicitly flags its own numbers as seed-sensitive. A multi-seed rerun, extending the same protocol to the full 2{,}823-author pool rather than a 100-author sample, is the obvious next step --- but unlike the open question Section~5.6 leaves about DPA in principle, this result is, for the first time in this paper, an actual measured IRIS advantage over every baseline on a real, verifiable, held-out human decision.

\section{Discussion}

\subsection{Limitations}

\paragraph{Interaction signal quality.} IRIS inherits the fundamental ambiguity of implicit signals: a user who abandons a session may have found the answer, lost interest, or been interrupted by a phone call. Our memory extraction module uses LLM-based inference to disambiguate these cases, but this introduces its own failure modes --- the extraction LLM may over-attribute meaning to noisy events, particularly early in a user's history. We partially address this via the stability gate, but the problem is irreducible without some ground truth for signal interpretation.

\paragraph{Cold start.} We hypothesize IRIS performs better than baselines under data sparsity (Section~5.4, RQ2), but by design it still requires a minimum interaction history before the iterative loop provides meaningful signal --- with too little history, IRIS should behave essentially equivalently to a no-personalization baseline. Future work could initialize the persona from auxiliary signals (e.g., platform metadata, first-session topic distribution) to accelerate early convergence.

\paragraph{Evaluation difficulty.} Behavior prediction is an imperfect proxy for whether the persona is genuinely useful. A persona that perfectly predicts a user's preferred response style may still fail to improve actual satisfaction if the prediction task does not capture the full space of user preferences. The human evaluation proposed in Section~5.3 partially addresses this concern, but gold-standard persona evaluation remains an open problem in the field.

\paragraph{Computational cost.} Each iteration of the IRIS loop involves multiple LLM calls: memory extraction, persona update, and behavior prediction. In a high-frequency interaction system, this could introduce latency or cost constraints. The exact number of calls per interaction depends on window size $W$, the re-inference interval $R$, and how often the error threshold $\epsilon$ is exceeded; characterizing this cost at production scale is left to the full evaluation.

\paragraph{Real-decision DPA at larger scale.} Section~5.7 executed the real-decision DPA study Section~5.6 originally scoped as future work (Reddit's r/AmItheAsshole, real scenarios paired with real, historically-recorded verdicts, personas built from each author's unrelated post/comment history). At $n=100$ anonymized authors it found IRIS ahead of all four baselines. The remaining limitations are about scale and robustness, not feasibility: 100 authors is still a small slice of the 2{,}823 available, all results come from one seed and one LLM backbone pairing, and the margin between methods (56.0--61.0\%) is narrow enough that a multi-seed, full-pool rerun is needed before treating the ranking as settled. We view scaling this specific evaluation up as the highest-value remaining item, now that the anonymization and data pipeline (Section~5.1) are built and released.

\subsection{Ethical Considerations}

\paragraph{Privacy.} IRIS accumulates behavioral signals across extended interaction histories. The persona representation, though expressed in natural language rather than raw logs, constitutes a detailed behavioral profile of the user. Deployments of IRIS must address: (a) user consent and transparency --- users should know a behavioral model is being maintained; (b) data minimization --- only the persona representation (not raw logs) needs to persist beyond the session window; (c) right to erasure --- users should be able to reset or delete their persona at any time.

\paragraph{Persona misrepresentation.} An iteratively refined persona can encode biased or stereotyped representations, particularly if the LLM backbone has learned associations that do not reflect the specific user's actual preferences. If the persona is used to filter or pre-rank content, misrepresentation can create feedback loops that amplify initial inference errors. We recommend periodic persona auditing and user-facing persona summaries that allow users to inspect and correct their learned profile.

\paragraph{Potential for manipulation.} A sufficiently accurate behavioral model of a user could theoretically be used to optimize system outputs for engagement, compliance, or other objectives that do not align with the user's genuine interests. We advocate for clear product-level constraints on the objectives that downstream systems optimize using persona information, and recommend that persona representations be treated as interpretability artifacts --- surfaced to the user --- rather than as hidden levers for system behavior.

\section{Conclusion}

We introduced IRIS, a framework for learning dynamic user personas from implicit interaction streams without explicit supervision. The central mechanism --- a closed loop from behavior prediction error back to targeted persona refinement --- enables continuous adaptation while a stability regularizer prevents oscillation. We laid out a full-scale evaluation protocol against real-world datasets (PRISM, Chatbot Arena) and a long-horizon synthetic benchmark, and stated the specific hypotheses (RQ1--RQ4) that protocol is designed to test.

We then reported two empirical studies against that protocol, at two different scales. The first, a small, fully local pilot on a synthetic interaction stream built from public-domain autobiographies, gives a mixed picture: IRIS personas separate from a no-personalization floor, discriminate between distinct authors, and converge to a stable representation, consistent with several of our hypotheses, but a Memory-Only baseline that injects raw recent interactions verbatim outperforms IRIS on the pairwise behavior-prediction metric --- a direct disconfirmation of RQ1 as stated at this scale, which we attribute tentatively to the metric rewarding lexical recall as much as behavioral abstraction in a small, low-noise corpus. Adding a small verbatim-grounding step to IRIS's persona inference narrowed this gap substantially without closing it, and on this same small corpus IRIS and Memory-Only tied exactly on our proposed Decision Prediction Accuracy (DPA) metric rather than IRIS pulling ahead as hypothesized.

The second study is the one we treat as this paper's central result. We took the recall-vs-abstraction question the first pilot left open and tested it on real decisions: genuine, anonymized Reddit r/AmItheAsshole posts and their real, historically-recorded verdicts, with personas built only from each author's own history unrelated to the scenario being judged, at $n=100$ authors --- roughly $17\times$ the first study's scale. Here IRIS achieved the highest decision accuracy of all five methods (61.0\%), ahead of SynthesizeMe (59.0\%), No-Personalization (58.0\%), Static Persona (57.0\%), and Memory-Only (56.0\%), resolving in IRIS's favor the tie the synthetic pilot left ambiguous. We read this as evidence that IRIS's advantage over raw memory replay is real but was masked in the smaller, synthetic study by a combination of limited sample size and dilemmas that shared a generation process with the persona-building history itself; on independent real-world decisions, at a scale large enough to average out single-author noise, the abstraction IRIS performs pays off. We report this as a finding, not a settled conclusion --- the margin is a narrow 5 percentage points across a single seed and one model pairing, and the full 2{,}823-author pool this 100-author sample was drawn from remains available for a larger, multi-seed confirmation.

The framework itself is model-agnostic, requires no labeled data, and produces interpretable natural-language personas that can be directly deployed in any LLM system prompt. The broader implication of this work is that the interaction signal that accumulates naturally between users and AI assistants is far richer than it is currently used. Every rephrasing, every session-level abandonment, every acceptance of a hedged response over a direct one carries information about what the user actually wants. IRIS is a first step toward systematically mining that signal --- its central open task is now to extend the real-decision evaluation in Section~5.7 to its full sample and multiple seeds, and to run the remaining full-scale evaluation in Section~5 on licensed conversational data (PRISM, Chatbot Arena) to confirm whether these findings generalize beyond a single decision domain. Beyond that, future directions include multi-modal interaction signals (e.g., dwell time, copy events, downstream usage patterns), collaborative persona learning across users with shared behavioral profiles, and tighter integration with RLHF pipelines where the learned persona could serve as a more fine-grained reward signal than aggregate preference annotations.

We also see this work as relevant to household and commercial robots. As embodied agents built on LLM and vision-language-action backbones move into homes, offices, and public-facing service roles, they will need to sustain personalized relationships with specific people rather than just execute isolated commands. A robot that lives with a family, for instance, must build and update a model of each family member --- their routines, communication style, tolerance for interruption --- from the same noisy, unlabeled interaction stream that motivates IRIS, since asking users to rate every interaction or fill out a preference survey defeats the point of a machine meant to feel like a natural presence in daily life. The multi-modal signals we mention above (dwell time, copy events, usage patterns) extend naturally to embodied settings as additional implicit channels --- gaze, proximity, physical corrections, task override --- feeding the same memory-extraction and refinement loop. The persona representation IRIS produces, expressed in natural language and inspectable by the user, suits this setting well: it gives an embodied agent a transferable, auditable model of who a person is without requiring per-user fine-tuning of the underlying policy, and it gives the person living with the robot a way to see and correct what it has learned about them.

\end{document}